\documentclass{article}

    \PassOptionsToPackage{numbers, compress}{natbib}

\usepackage[final]{ARLET_2024}

\usepackage[utf8]{inputenc} 
\usepackage[T1]{fontenc}    
\usepackage{hyperref}       
\usepackage{url}            
\usepackage{booktabs}       
\usepackage{amsfonts}       
\usepackage{nicefrac}       
\usepackage{microtype}      
\usepackage{xcolor}         

\usepackage{graphics} 
\usepackage{epsfig} 
\usepackage{mathptmx} 
\usepackage{times} 
\usepackage{amsmath} 
\usepackage{amssymb}  
\usepackage{subfigure}

\title{COSBO: Conservative Offline Simulation-Based Policy Optimization}

\author{%
  Eshagh Kargar \\
  \texttt{eshagh.kargar@aalto.fi} \\
  \And
  Ville Kyrki \\
  \texttt{ville.kyrki@aalto.fi} \\
}

\begin{document}

\maketitle

\begin{abstract}
Offline reinforcement learning allows training reinforcement learning models on data from live deployments. However, it is limited to choosing the best combination of behaviors present in the training data. In contrast, simulation environments attempting to replicate the live environment can be used instead of the live data, yet this approach is limited by the simulation-to-reality gap, resulting in a bias. 
In an attempt to get the best of both worlds, we propose a method that combines an imperfect simulation environment with data from the target environment, to train an offline reinforcement learning policy. 
Our experiments demonstrate that the proposed method outperforms state-of-the-art approaches CQL, MOPO, and COMBO, especially in scenarios with diverse and challenging dynamics, and demonstrates robust behavior across a variety of experimental conditions. The results highlight that using simulator-generated data can effectively enhance offline policy learning despite the sim-to-real gap, when direct interaction with the real-world is not possible.
\end{abstract}

\section{Introduction}

Offline reinforcement learning (offline RL)~\cite{lange2012batch, kumar2020conservative} is defined as the task of learning the best possible policy from a fixed and pre-collected dataset without having interaction with the world. 
This leads to a setting for data reuse and safe policy learning applicable to robotics~\cite{singh2020cog, yu2021combo, rafailov2021offline}, autonomous driving~\cite{fu2020d4rl, graves2020learning}, and healthcare~\cite{tang2021model}.

Directly applying online RL algorithms to offline RL settings produces poor results~\cite{kidambi2020morel, fujimoto2019off, yu2020mopo} because of the distribution shift between the offline dataset and the learned policy. So it needs designing algorithms specialized for offline RL~\cite{yu2021combo}. 
Model-based algorithms, which learn a dynamics model from logged experience and perform some sort of pessimistic planning under the learned model, have emerged as a promising paradigm for offline reinforcement learning (offline RL).
Some algorithms such as MOPO~\cite{yu2020mopo} and MOReL~\cite{kidambi2020morel} try to do explicit uncertainty quantification for incorporating pessimism. Also, there are some other methods such as COMBO~\cite{yu2021combo}, which try to regularize the value function on out-of-support state-action tuples generated via roll-outs under the learned model to estimate a conservative value function without any explicit uncertainty estimation. However, learning dynamic models via deep learning can be difficult and unreliable and requires significant human expertise and hyper-parameter tuning to avoid collapsing~\cite{zhang2021importance}. 


On the other hand, for some tasks such as autonomous driving and robotics, in addition to a fixed and pre-collected dataset, we can have access to a simulator that has a dynamic mismatch. Using the simulation-generated data can help to learn a better policy~\cite{jiang2020offline}. In this work, our goal is to propose a new algorithm that retains the benefits of offline RL and uses simulation-generated data while removing the need for model learning.


\begin{figure}[t]
\includegraphics[width=\columnwidth]{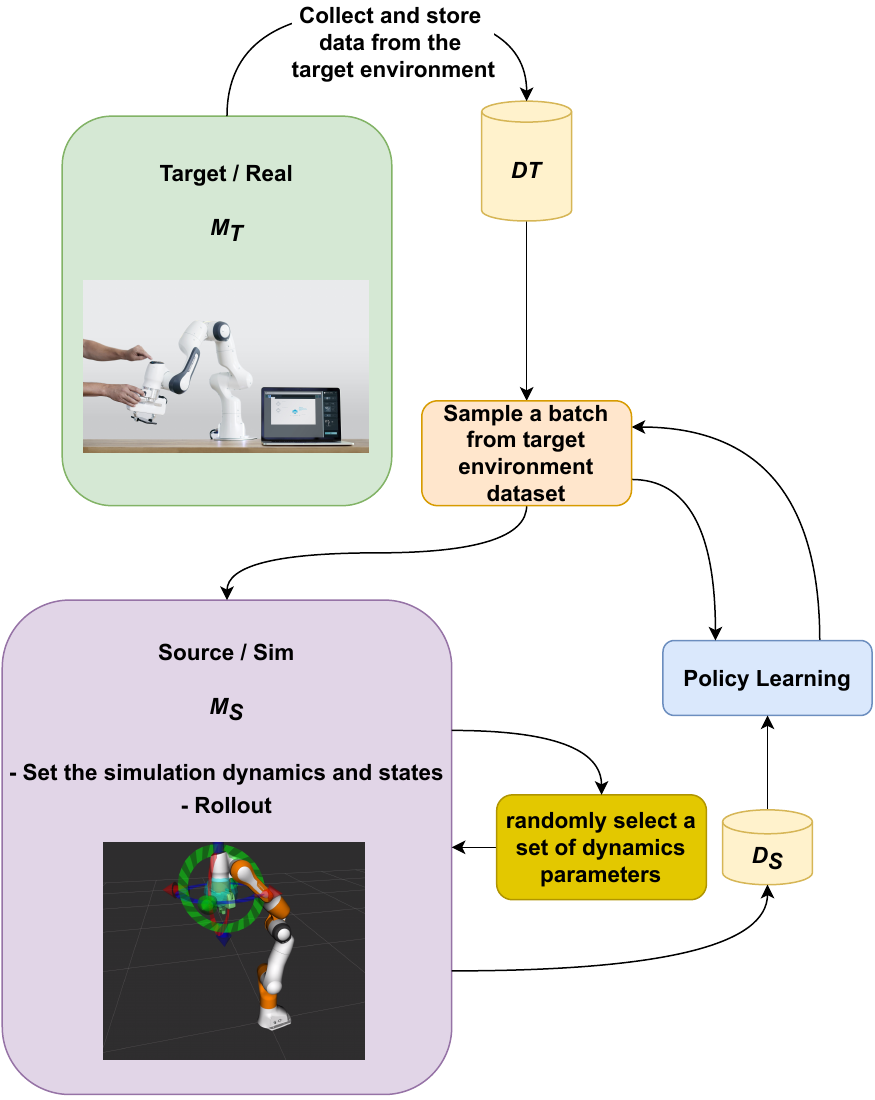}
\centering
\caption{The proposed COSBO framework. }
\label{fig:cosbo}
\end{figure}

Our main contribution is developing a new simulation-based offline RL algorithm, COSBO, that regularizes the value function on out-of-support state-action tuples generated via rollouts under the simulation environments with different dynamics. 
This results in a conservative estimate of the value function for out-of-support state-action tuples, without requiring explicit uncertainty estimation and model learning.
This allows us to use simulation environments with different dynamics to generate data and estimate a more generalized value function. Our proposed method can also be seen as a way to use COMBO for sim-to-real tasks.
Unlimited access to a simulator with close dynamics to the real environment can lead to a value function estimation that models the true value function on out-of-distribution but close data points better than COMBO and other model-free and model-based algorithms. 
The closest work to ours is COMBO. However, in contrast to COMBO, COSBO does not need to learn a model to generate data. It uses a simulator with different but close dynamics to the target environment to generate data. COSBO uses both offline data and simulation-generated data to update its Q-function. Note that COSBO does not mix these two sources of data but proposes a new way which will be detailed in the next sections. 
COSBO learns a tighter lower bound for the true value function which leads to higher rewarding policies compared to COMBO and other baselines.
Lastly, in our experiments, we find that COSBO outperforms state-of-the-art methods in several domains from D4RL benchmark~\cite{fu2020d4rl} and a real-world robotic environment.

The rest of this paper is organized as follows. 
The review of related works in Section \ref{related_works} demonstrates that current offline RL algorithms, both model-free and model-based, have distinct advantages and limitations.
In Section \ref{background}, we provide the required background in Markov Decision Process and Offline RL. 
The problem definition and the proposed method are described in Section \ref{method}, with emphasis on how COSBO leverages simulation data to enhance learning and addresses the limitations of existing methods. 
Then, Section \ref{experiments} presents empirical evaluation, comparing COSBO with state-of-the-art algorithms and analyzing its performance under various dynamics mismatches. 
Finally, in Section \ref{conclusion}, we conclude by summarizing our findings, highlighting the effectiveness of COSBO in improving offline RL performance, and discussing its potential applications and limitations in real-world scenarios.

\section{Related Work}
\label{related_works}


Offline RL algorithms aim to train RL agents using only pre-collected datasets. To tackle the issue of state-conditional action distribution shift, previous methods either (a) explicitly constrain the policy to remain close to the behavior policy or (b) train pessimistic value functions.


One approach in offline RL algorithm design is to incorporate conservatism or regularization into the learning process. Model-free offline RL algorithms \cite{fujimoto2019off, kumar2019stabilizing, wu2019behavior, jaques2019way, kumar2020conservative} directly integrate conservatism into policy or value function training without needing a dynamics model. However, these algorithms only learn from the states in the offline dataset, which can result in overly conservative behavior.


In contrast, model-based algorithms \cite{kidambi2020morel,yu2020mopo} learn a pessimistic dynamics model that leads to a conservative value function estimate. Model-based RL has shown an early advantage in data efficiency over value-function-based RL by explicitly learning a transition model \cite{sutton1990integrated}. This advantage has been repeatedly demonstrated \cite{deisenroth2011pilco, kurutach2018model, nagabandi2018neural}. However, these algorithms rely on accurate uncertainty quantification of the learned dynamics model, which can be challenging for complex datasets or deep networks. Additionally, these methods do not adapt the uncertainty estimates as the policy and value function evolve during learning. Unlike MOReL \cite{kidambi2020morel}, which uses a hard threshold on uncertainty to avoid drifting to unknown states, MOPO uses a soft reward penalty to handle uncertainty. This soft penalty allows the policy to take some risky actions and then return to safer regions near the behavior distribution without being terminated. COMBO, on the other hand, removes the need for uncertainty quantification.



Jiang et al. \cite{Jiang2020OfflineIL} use simulator states as goals based on expert states, recovering actions through horizon-adaptive inverse dynamics in an offline imitation learning setting. While the state-dependent discriminator helps focus on the expert distribution, it requires thorough training. Unlike offline imitation learning, which can learn better policies from data with rewards, PerSim \cite{Agarwal2021PerSimDO} learns personalized policies simultaneously across heterogeneous sources (agents) but needs low-rank latent factor representations. Despite its promising performance on standard RL benchmarks, PerSim cannot yet be used as an out-of-the-box solution for critical real-world problems without a rigorous validation framework.

\section{Background}
\label{background}

In this section, we will review the required theoretical concepts behind the proposed method, including Markov Decision Process, reinforcement learning, offline RL, model-based and model-free RL, and COMBO. 

\subsection{Markov Decision Process and Offline RL}
We consider the standard Markov Decision Process (MDP) $M = (S, A, T, r, \mu_0, \gamma)$ where $S$ and $A$ denote the state space and action space, respectively, $T(s'|s,a)$ the transition function, $r(s,a)$ the reward function, $\mu_0$ the initial state distribution, and $\gamma \in (0, 1)$ the discount factor. 
We denote the discounted state visitation distribution of a policy $\pi$ using $d^\pi_M(s) := (1 - \gamma) \sum_{t=0} ^ \infty \gamma^t P(s_t = s|\pi)$, where $P(s_t = s|\pi)$ is the probability of reaching state $s$ at time $t$ by rolling out $\pi$ in $M$.
Also, we denote the state-action visitation distribution with $d_M^\pi(s, a) := d_M^\pi(s)\pi(a|s)$.
Reinforcement learning aims to find an optimal control policy $\pi:S\rightarrow A$ from states to actions that maximizes total expected future rewards
\begin{equation}\label{rl_reward}
     R(\pi) = E_{\pi} \Bigg[ \sum_{t} \gamma^t r(s_t, a_t) \Bigg].
\end{equation}
In the offline RL setting, which can be defined as a data-driven formulation for reinforcement learning settings~\cite{levine2020offline}, the algorithm only has access to a fixed and pre-collected dataset $D = {(s, a, r, s')}$ which is collected by one or a mixture of behavior policies $\pi_\beta$. In this formulation, the agent cannot interact with the environment to collect additional data. 

The approaches to solving MDPs can be categorized into two classes: model-free and model-based.
Model-free RL approaches use dynamic programming and actor-critic schemes~\cite{sutton2018reinforcement, bertsekas1995neuro} and do not need to learn a dynamics model. 

On the other hand, model-based methods try to learn an MDP $\hat{M} = (S, A, \hat{T}, r, \mu_0, \gamma)$ which uses the learned transition dynamics $\hat{T}$ instead of the true transition function $T$. 
In this paper, we assume the reward function $r$ is known; however, it can be learned from data if it is unknown.
Similarly, the initial state distribution $\mu_0$ can either be learned if it is unknown too.

\subsection{COMBO}
It is difficult to quantify uncertainty in all environments, especially complex ones, which is one of the main drawbacks of model-based algorithms before COMBO. COMBO proposes an offline RL algorithm that optimizes a lower bound for the policy performance without explicitly estimating uncertainty~\cite{yu2021combo}. This approach combines model-free and model-based elements to leverage the strengths of both methods. By learning a dynamics model and using it to generate synthetic data, COMBO mitigates the conservativeness of model-free algorithms and avoids the challenges of uncertainty quantification in model-based approaches. This results in a more robust and efficient learning process that can handle complex environments effectively.

\section{Method}
\label{method}

Model-based RL algorithms have the major challenge that learning dynamic models through deep learning can be complicated and unreliable, requiring significant human expertise to avoid collapsing~\cite{zhang2021importance}. In this work, we propose a solution: an offline RL algorithm that uses simulation data in a new way to optimize a lower bound on policy performance without learning a dynamic model.

\subsection{Conservative Policy Evaluation}
With a policy of $\pi$, an offline dataset of $D$, and a simulator $S$ of the environment in which the offline dataset was collected, we intend to learn a conservative estimate of $Q^\pi$. 
In order to accomplish this, Q-values are penalized on state-action pairs drawn from a distribution that is more likely to be out-of-support while Q-values on trustworthy state-action pairs are pushed up. Implementation is achieved by recursion as follows:
\begin{equation}
    \begin{split}
        \hat{Q}^{k+1} \leftarrow \arg\min\limits_{Q} \beta \Big(E_{s,a \sim \rho (s,a)} [Q(s,a)] - E_{s,a \sim D} [Q(s,a)] \Big) \\ 
        + \frac{1}{2} E_{s,a,s' \sim d_f} \Bigg[\Big(Q(s,a) - \hat{B}^\pi \hat{Q}^k(s,a) \Big)^2 \Bigg] 
    \end{split}
\end{equation}
where $\hat{B}^\pi Q(s,a) := r(s,a) + \gamma Q(s', a')$ associated with a single transition $(s, a, s')$ and $a' \sim \pi(.|s')$. $\rho(s,a)$ and $d_f$ are sampling distributions, chosen as
\begin{equation}
    \rho(s,a) = d^\pi_S (s) \pi (a|s)
\end{equation}
\begin{equation}
    d_f(s,a) := f d(s,a) + (1 - f) d^\mu_S (s,a)
\end{equation}
where $d^\pi_S (s)$ is the discounted marginal state distribution when executing $\pi$ in the simulator $S$. Samples from $d^\pi_S (s)$ can be obtained by rolling out $\pi$ in $S$.
Also, $d_f$ is an $f$-interpolation between the offline dataset and the generated data from rollouts in the simulator using the rollout distribution, $\mu (.|s)$.

Using a mixture of the simulator and real data, we update Q-values with Bellman backup and push down (or estimate conservatively) Q-values on state-action pairs resulting from simulator rollouts and push-up Q-values on the offline state-action pairs. 
COSBO, in contrast to COMBO and other model-based algorithms, does not learn a dynamics model to generate data but uses a simulator instead. 

\subsection{Policy Improvement}
The policy can be improved using the learned conservative critic $\hat{Q}^\pi$ as follows:
\begin{equation}
    \pi' \leftarrow \arg\max \limits_{\pi} E_{s \sim \rho, a \sim \pi(.|s)} \Big[\hat{Q}^\pi (s,a) \Big]
\end{equation}
where $\rho(s)$ is the state marginal of $\rho(s,a)$. The $argmax$ can be approximated with a few steps of gradient descent~\cite{yu2021combo} when policies are parametrized with neural networks. Additionally, to prevent the policy from becoming degenerate an entropy regularization can be used~\cite{haarnoja2018soft}.

\section{Experiments}
\label{experiments}

We conducted a series of experiments to address the following research questions:

\begin{enumerate}
    \item How does the proposed method compare against state-of-the-art offline RL techniques, when using the original dataset (excluding data generated by the simulator)? \label{q1}
    
    \item What is the best way to utilize simulation data? Should it be concatenated with offline data from the target environment to train COMBO, or should our method be used to incorporate this data instead of relying on generated data from the learned model? \label{q2}
    
    \item How does changing the simulation dynamics affect the performance, and how large can the gap between simulation and the target environment be to yield useful results? \label{q3}
    
    
\end{enumerate}

\subsection{Environments}
To evaluate our proposed method, we tested it on two simulation tasks. We used environments with standard dynamics as the target (real) environments and modified the dynamics to create simulation environments with similar but slightly different dynamics. Specifically, we considered the Hopper and Walker2d tasks from the MuJoCo simulation suite.

\subsection{Implementation Details}
Our implementation is largely based on COMBO~\cite{yu2021combo}. Although there is no official implementation of COMBO, we utilized an open-source version provided by Polixir~\footnote{https://agit.ai/Polixir/OfflineRL/src/branch/master} and adapted it to develop our method. The primary difference is that we do not learn any transition model; instead, we use a simulator with dynamics close to the main environment from which the offline dataset was collected. The rest of the implementation details align with those in the COMBO paper~\cite{yu2021combo}. For a fair comparison, we used all the baselines from this repository.

\subsection{Simulation Dynamics}
This section describes the dynamics used in the simulator to generate data. For our experiments, we used datasets from D4RL~\cite{fu2020d4rl} as the target environment, representing a fixed real-world dataset. We then employed a simulator with varying dynamics to simulate the mismatch between the offline dataset from the target environment and an imperfect simulator. We sampled state-action pairs from the offline dataset $D$, set the simulator to those states, and applied the actions. For each sample, we randomly selected one of the following dynamics:

\begin{itemize}
    \item Light: the mass is half of the normal mass in $D$.
    \item Heavy: the mass is $1.5\times$ the normal mass in $D$.
    \item Short: the length is half of the normal length in $D$.
    \item Long: the length is $1.5\times$ the normal length in $D$.
\end{itemize}

\subsection{Comparison to Baselines}
\begin{figure}[t!]
    \centering
    \subfigure[Hopper: D4RL dataset only]{
        \includegraphics[width=0.45\textwidth]{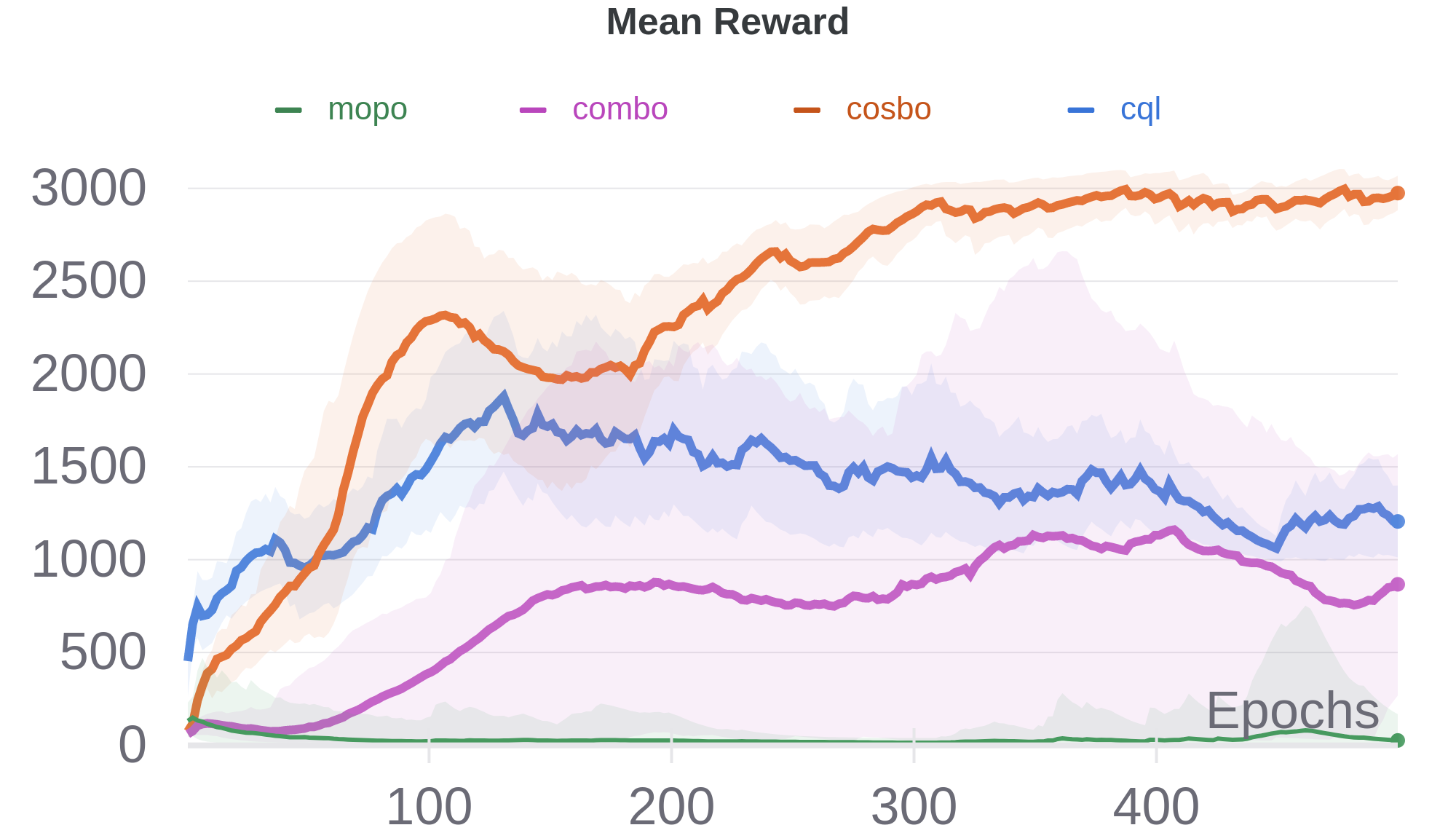}
    }
    \subfigure[Walker2d: D4RL dataset only]{
        \includegraphics[width=0.45\textwidth]{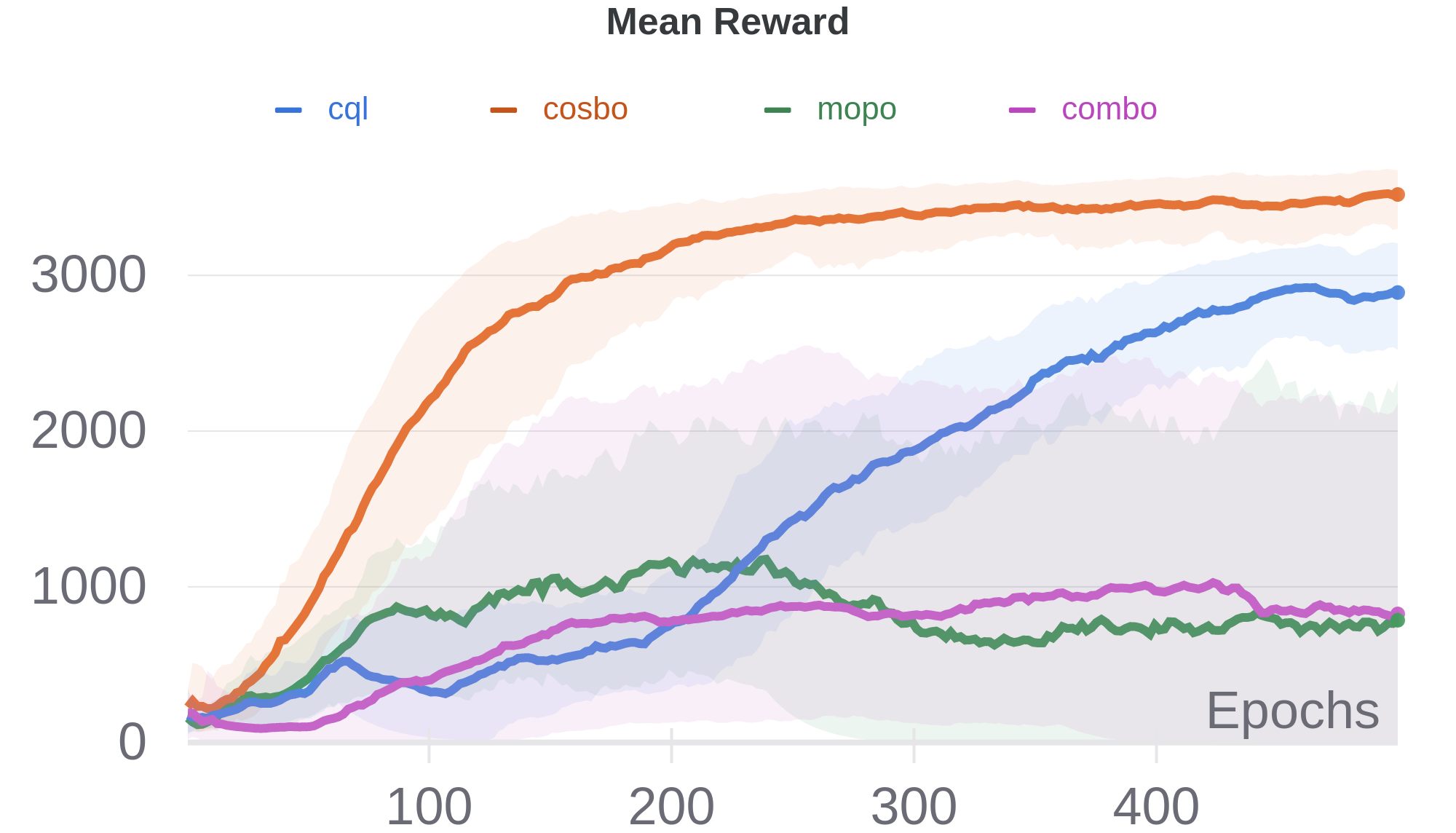}
    }
    \caption{Comparison to baselines in the Hopper and the Walker2d environments while using the D4RL dataset only.}
    \label{fig:hopper_walker2d_sota_d4rl}
\end{figure}

In response to question Q~\ref{q1}, we compared our method with CQL~\cite{kumar2020conservative}, MOPO~\cite{yu2020mopo}, and COMBO~\cite{yu2021combo}, which are state-of-the-art model-free and model-based algorithms, using only the dataset from the target environment. In this experiment, the baseline algorithms utilized the target environment dataset, specifically hopper-medium. Model-based methods like MOPO and COMBO learn a dynamics model to generate data for their algorithms, whereas CQL, being a model-free algorithm, does not generate additional data. Our method, COSBO, uses a simulator with similar but different dynamics. COSBO employs $(s, a)$ pairs from $D$ and generates data by rolling out from state $s$ and applying action $a$. Since the simulator's dynamics differ, it transitions to a new state $s''$ that is not present in $D$. Essentially, if $(s, a, s')$ is a sample from $D$, the generated data would be $(s, a, s'')$. Fig~\ref{fig:hopper_walker2d_sota_d4rl} shows the results.

\begin{figure}[t!]
    \centering
    \subfigure[Hopper: D4RL+simulation data (medium change)]{
        \includegraphics[width=0.45\textwidth]{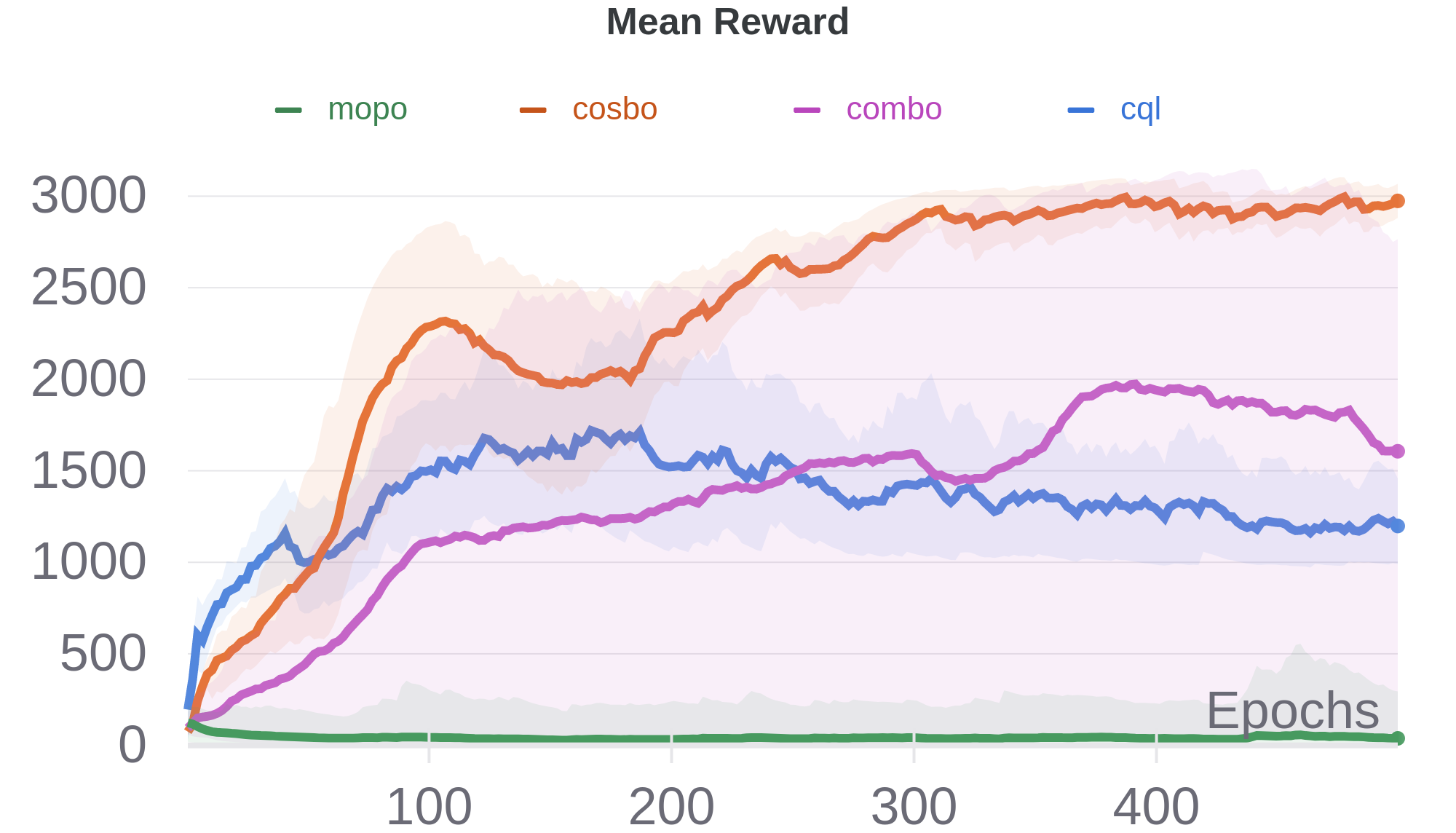}
    }
    \subfigure[Walker2d: D4RL+simulation data (medium change)]{
        \includegraphics[width=0.45\textwidth]{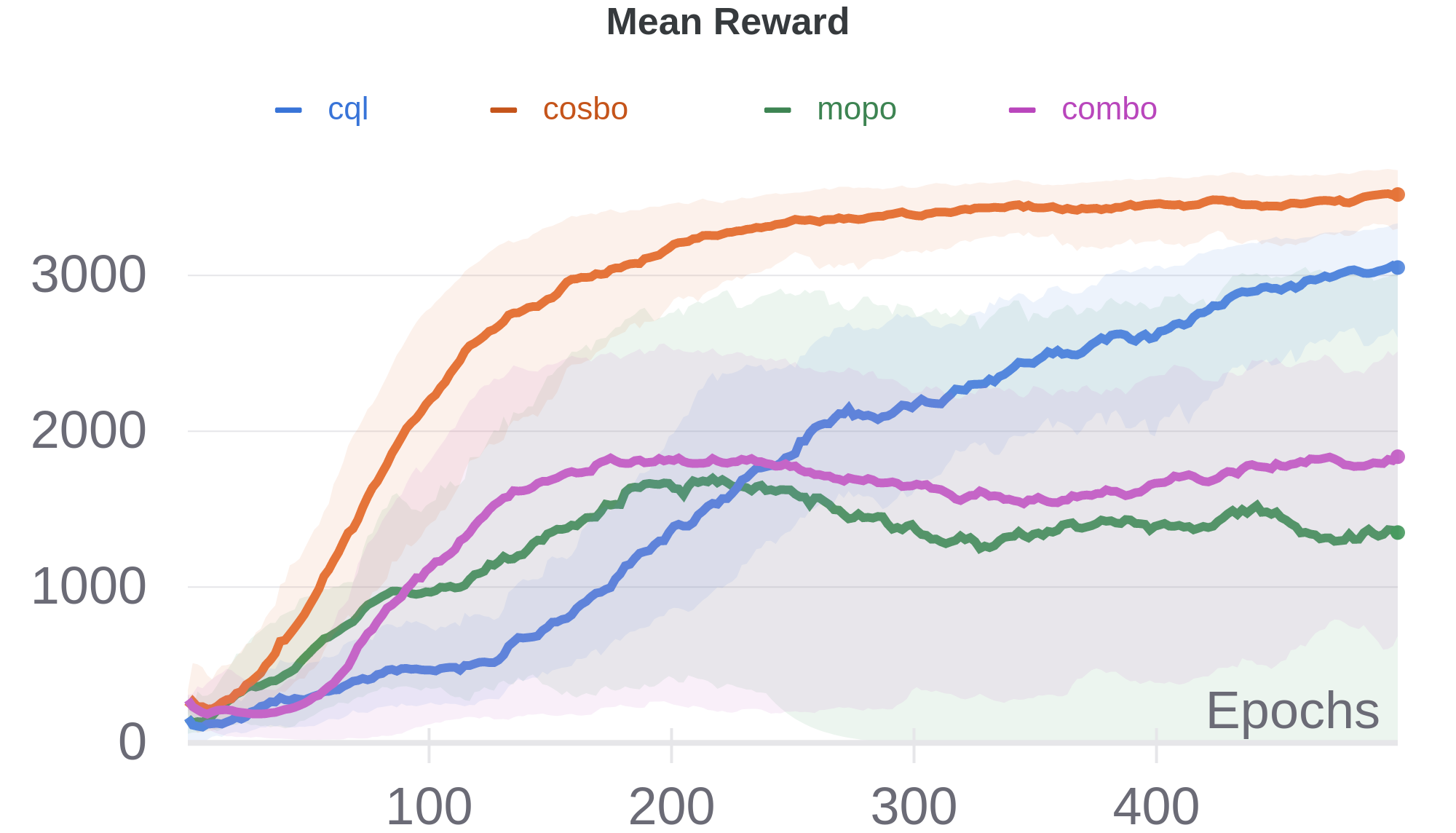}
    }
    \caption{Comparison to baselines in the Hopper and the Walker2d environments while using the D4RL+simulation data (medium change).}
    \label{fig:hopper_walker2d_sota_medium}
\end{figure}

To address question Q~\ref{q2}, we investigated whether the performance improvement by COSBO is due to the newly generated data from the simulation or the method itself. We trained baselines using the concatenation of simulation-generated data and the main target dataset and compared them with our proposed method. This means the other methods used the same dataset as our method but with a different approach. As seen in Fig~\ref{fig:hopper_walker2d_sota_medium}, baselines could not match COSBO's performance even with the same dataset. This highlights the effectiveness of our method in utilizing simulation data, rather than merely concatenating it with the target dataset. Notably, while CQL's performance remained unchanged, COMBO's performance improved. An interesting observation with COMBO was the variability in performance due to different random seeds, model weight initialization, and hyperparameters, indicating the sensitivity of model-based methods to these factors. In contrast, our algorithm demonstrated much more stable performance.

In addressing question Q~\ref{q3}, we explored how the performance is affected by the degree of difference between the simulator dynamics and the target environment. We generated data by altering the dynamics in the environment, considering several cases with "Very" and "Extreme" dynamic changes as shown in Table~\ref{tab:dynamics}. The datasets for the Very/Extreme cases included four dynamic variations: very/extreme light, very/extreme heavy, very/extreme short, and very/extreme long. In each scenario, we changed only one dynamic at a time while keeping all others constant. For example, in the very light version, the mass was $0.3$ times that of the target environment, with no change in length, and in the very long version, the length was $2$ times that of the target environment.

\begin{table}[t!]
    \centering
    \caption{Different dynamics for the simulator.}
    \label{tab:dynamics}
    \begin{tabular}{|*{5}{c|}}\toprule
                & Light             & Heavy             & Short               & Long                \\ \midrule
     Very       & $0.3 \times mass$ & $2 \times mass$   & $0.3 \times length$ & $2 \times length$   \\
     Extreme    & $0.1 \times mass$ & $3 \times mass$   & $0.1 \times length$ & $3 \times length$    \\  
     \bottomrule
    \end{tabular}
\end{table}

\begin{figure}[t!]
    \centering
    \subfigure[Hopper: D4RL+simulation data (very change)]{
        \includegraphics[width=0.45\textwidth]{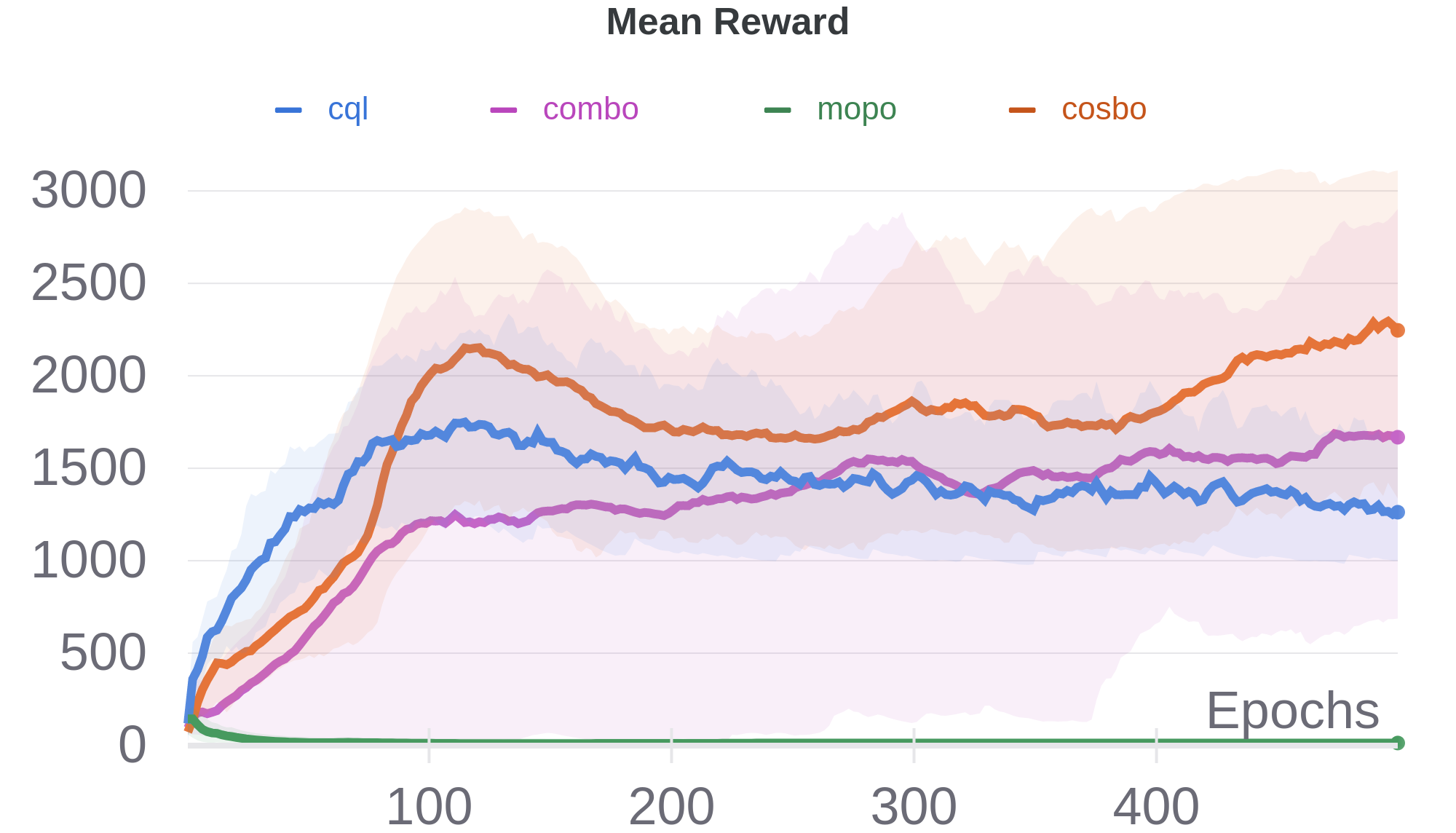}
    }
    \subfigure[Hopper: D4RL+simulation data (extreme change)]{
        \includegraphics[width=0.45\textwidth]{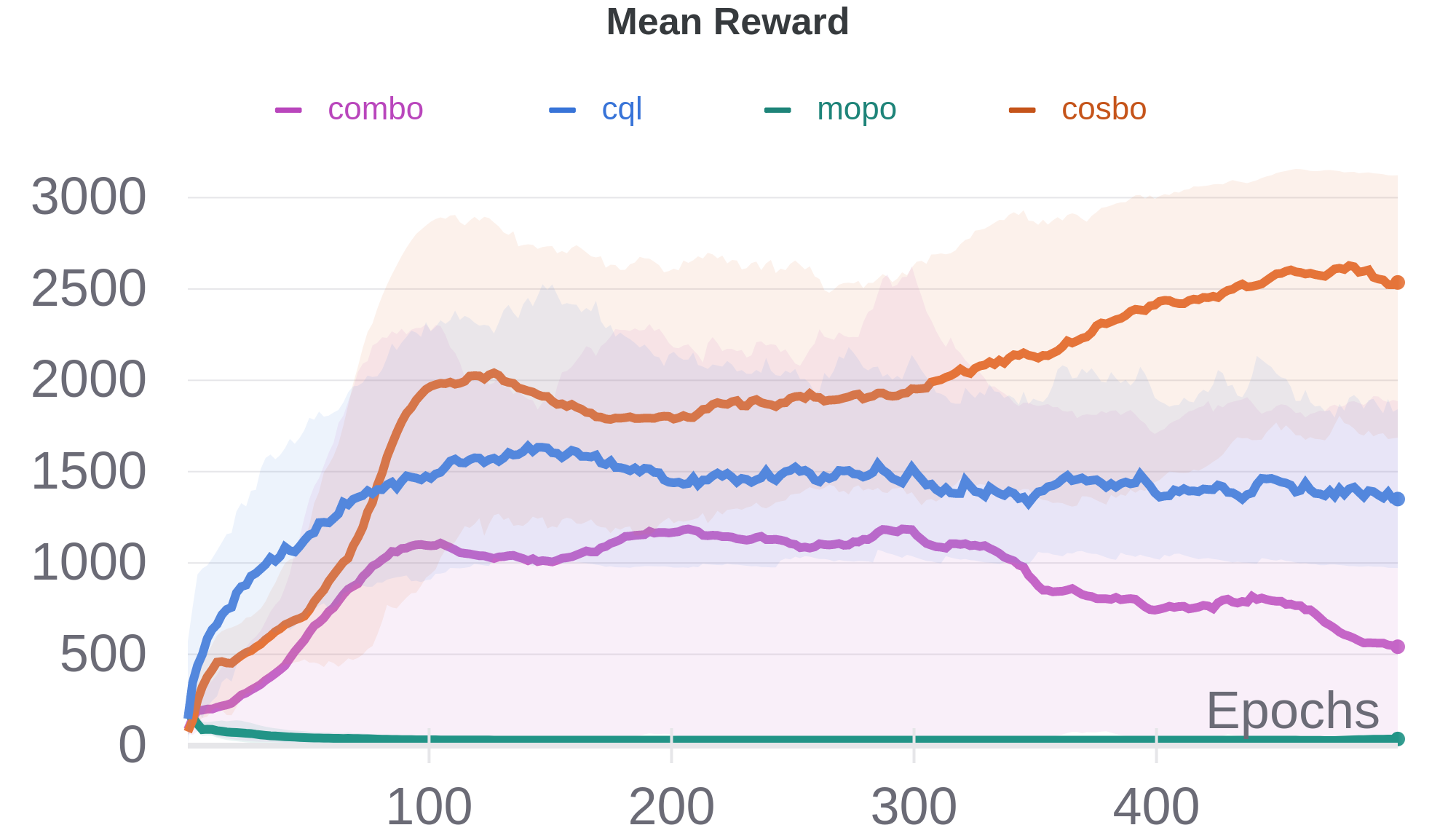}
    }
    \caption{Comparison to baselines in the Hopper environment with varying dynamics. (a) Using D4RL+simulation data (very change). (b) Using D4RL+simulation data (extreme change).}
    \label{fig:hopper_sota_mismatch}
\end{figure}

\begin{figure}[t!]
    \centering
    \subfigure[Walker2d: D4RL+simulation data (very change)]{
        \includegraphics[width=0.45\textwidth]{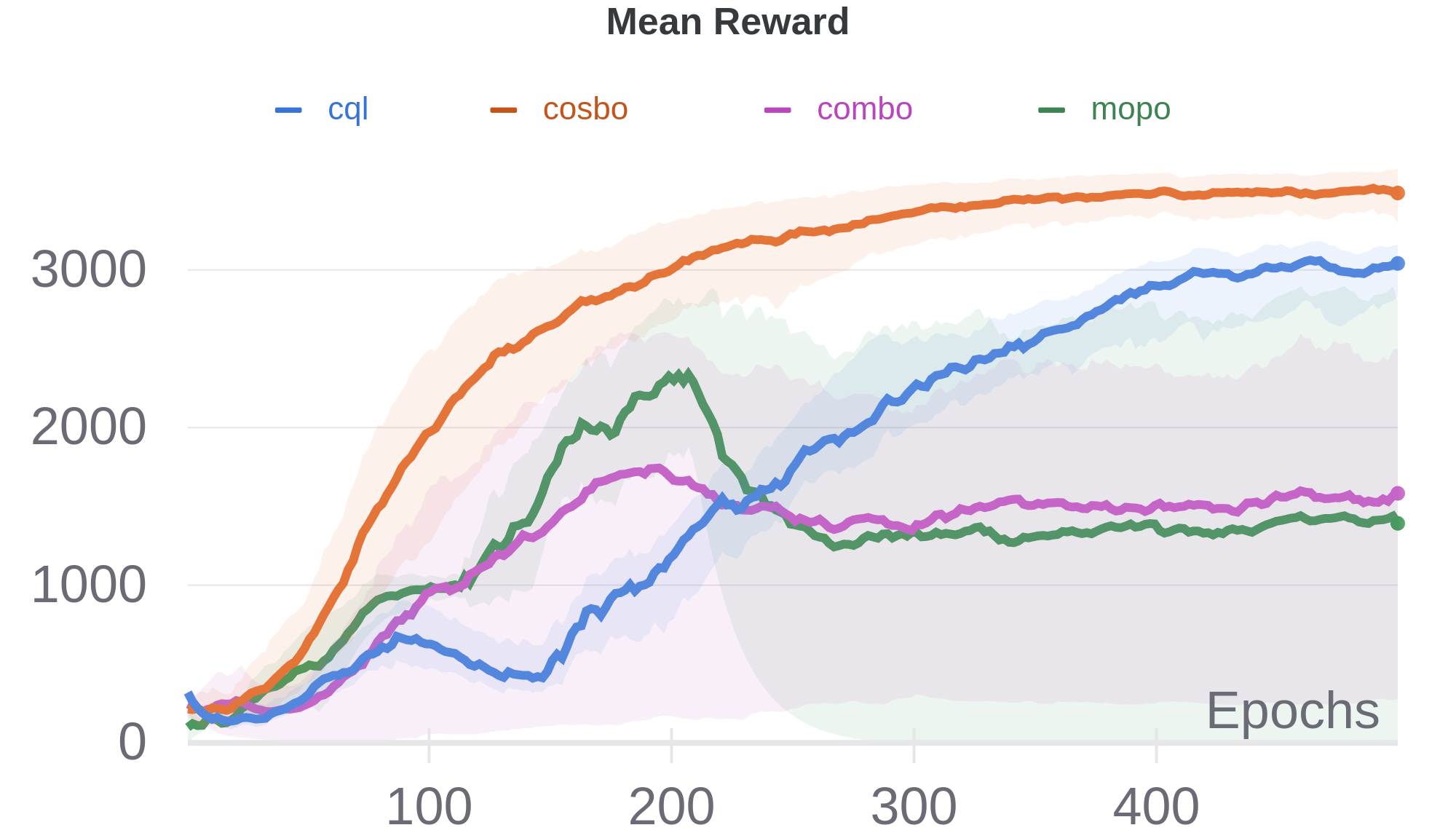}
    }
    \subfigure[Walker2d: D4RL+simulation data (extreme change)]{
        \includegraphics[width=0.45\textwidth]{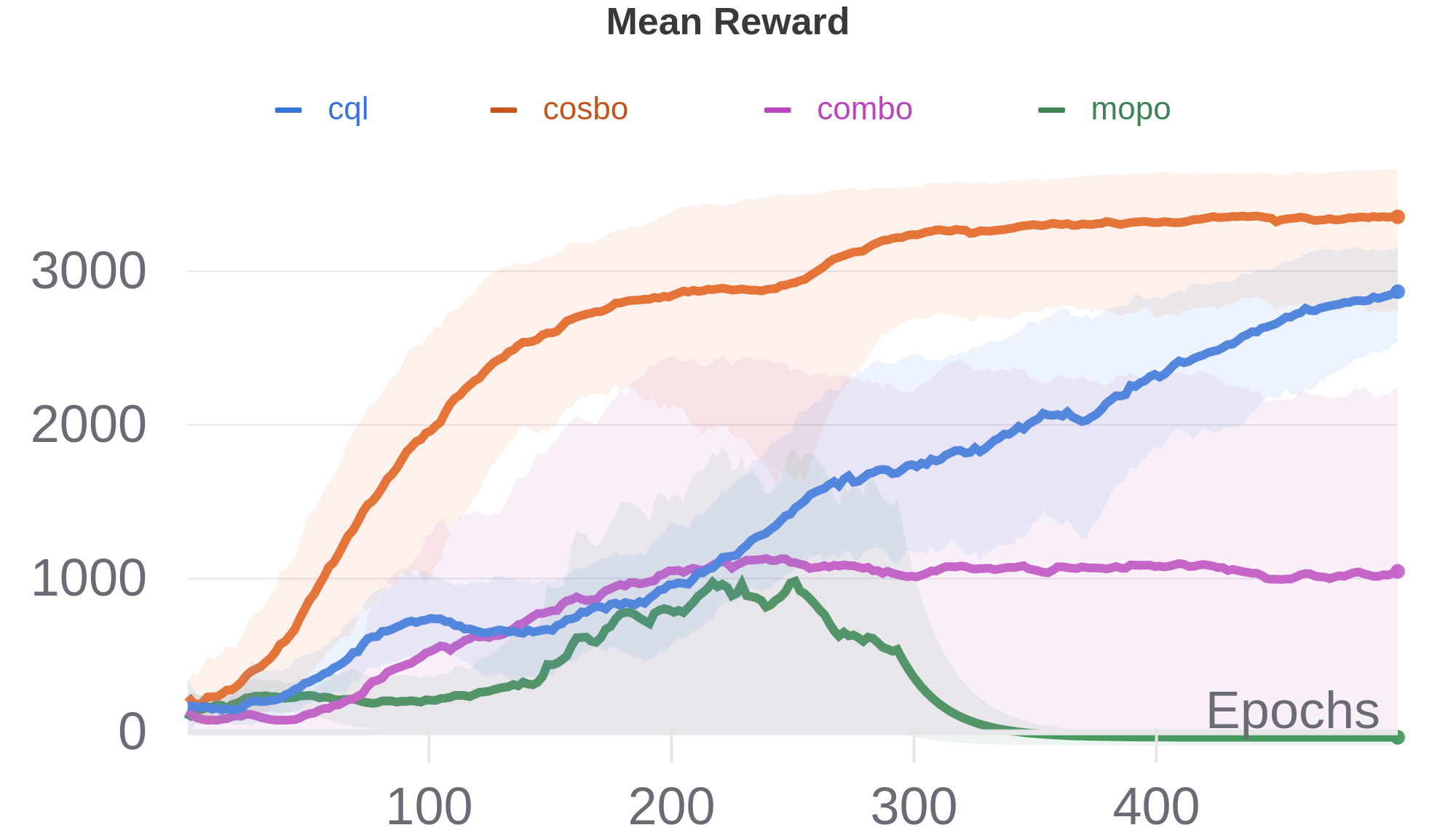}
    }
    \caption{Comparison to baselines in the Walker2d environment with varying dynamics. (a) Using D4RL+simulation data (very change). (b) Using D4RL+simulation data (extreme change).}
    \label{fig:walker2d_sota_mismatch}
\end{figure}

Figures \ref{fig:hopper_sota_mismatch} and \ref{fig:walker2d_sota_mismatch} display the results for the Hopper and Walker2d environments, respectively, under varying dynamics. These experiments demonstrate COSBO's robustness across different dynamic scenarios, maintaining superior performance even with significant dynamic mismatches.

\section{Conclusion}
\label{conclusion}

In this paper, we introduced COSBO, a novel offline RL algorithm that leverages simulation data to enhance learning. Our approach addresses the limitations of both model-free and model-based methods by using a simulator with similar but different dynamics to the target environment, thus generating useful and diverse data for training. 

We conducted extensive experiments to evaluate the effectiveness of COSBO, comparing it to state-of-the-art algorithms such as CQL, MOPO, and COMBO. The results demonstrated that COSBO outperforms these baselines, particularly in scenarios with diverse and challenging dynamics. Our method not only improves performance but also shows greater stability and robustness across different experimental conditions.

Furthermore, we explored the impact of varying dynamics in the simulation environment and found that COSBO maintains superior performance even with significant discrepancies between the simulator and the target environment. This highlights the potential of our approach to be applied in real-world settings where perfect modeling of the environment is infeasible.

In summary, COSBO provides a robust and efficient solution for offline RL by effectively integrating simulation data, thereby overcoming the conservatism of model-free methods and the complexity of uncertainty quantification in model-based methods. Future work could explore extending this approach to more complex tasks and further improving the integration of simulation data to enhance generalization and performance.








\bibliographystyle{plainnat}
\bibliography{ref}












\end{document}